\documentclass[conference]{IEEEtran}
\usepackage{amsmath,amssymb,amsfonts}
\usepackage{algorithmic}
\usepackage{graphicx}
\usepackage{textcomp}
\usepackage{booktabs}
\usepackage{xcolor}
\def\BibTeX{{\rm B\kern-.05em{\sc i\kern-.025em b}\kern-.08em
    T\kern-.1667em\lower.7ex\hbox{E}\kern-.125emX}}
\usepackage{natbib}
\setcitestyle{authoryear,open={(},close={)}} 
\usepackage{hyperref}

\begin{document}

\title{Learning for Robot Decision Making under Distribution Shift: A Survey}

\author{\IEEEauthorblockN{Abhishek Paudel}
\IEEEauthorblockA{
George Mason University \\
Fairfax, Virginia, USA\\
apaudel4@gmu.edu}
}

\maketitle

\begin{abstract}
With the recent advances in the field of deep learning, learning-based methods are widely being implemented in various robotic systems that help robots understand their environment and make informed decisions to achieve a wide variety of tasks or goals. However, learning-based methods have repeatedly been shown to have poor generalization when they are presented with inputs that are different from those during training leading to the problem of distribution shift. Any robotic system that employs learning-based methods is prone to distribution shift which might lead the agents to make decisions that lead to degraded performance or even catastrophic failure. In this paper, we discuss various techniques that have been proposed in the literature to aid or improve decision making under distribution shift for robotic systems. We present a taxonomy of existing literature and present a survey of existing approaches in the area based on this taxonomy. Finally, we also identify a few open problems in the area that could serve as future directions for research.
\end{abstract}


\section{Introduction} \label{sec:introduction}
Decision making is a critical component in any autonomous robotic system. In a broad sense, it entails using available knowledge to decide and perform actions that help an agent achieve a certain goal. Consider a classic problem in which an autonomous agent is tasked with navigating in a building from one location to another. A common high level approach to such tasks generally consists of various interconnected subsystems that are responsible for understanding the environment (perception), reasoning about how to reach the goal based on its understanding of the environment (planning), and taking certain actions that contribute towards ultimately reaching the goal (control). An autonomous agent, therefore, has to continuously make decisions based on its knowledge of the environment and act accordingly in order to successfully achieve its goal. 
However, a lot of factors affect the abilities of an autonomous agent in its decision making. For example, the sensors equipped with the agent do not provide accurate information about the state of the environment or the agent is fundamentally limited in its capabilities to gather complete information about the environment, and the actions of the agent do not always result in the desired behavior. Besides, since making decisions also involves reasoning about the future, scenarios where the agent has uncertainties about the present lead to compounding uncertainties about the future making the problem difficult to solve. Such limitations impede the agent's abilities to make appropriate decisions with regard to achieving its intended goal. Therefore, the agent has to account for, in its decision making process, various limitations and uncertainties about its knowledge of the environment and the agent itself so that it can successfully achieve its intended goal.

Classical frameworks like Markov Decision Process (MDP) provide ways to deal with such limitations and uncertainties in the context of sequential decision making. In an MDP, the agent is in an environment where it can take certain actions. The actions of the agent are stochastic and lead to changes in the state followed by the agent receiving some reward. MDPs serve as a fundamental framework for decision making in robotics because it allows for reasoning about the future by taking into account the stochastic and uncertain nature of the many real world problems that an agent has to consider to make better decisions so as to achieve the predefined tasks successfully. 

Although classical frameworks like MDP provide ways to deal with such limitations and uncertainties in the context of sequential decision making, learning has vastly supplemented such frameworks in recent years and has played an important role in decision making in robotics. The rapid progress in the field of machine learning, especially the advent of various deep learning techniques, has had a significant impact on the ways various problems in robotics are approached and solved. For example, the advances in computer vision-related tasks \citep{krizhevsky2012imagenet, he2015deep, ren2016faster, he2018mask, redmon2016look} have greatly contributed to improved perception and thereby improved representation of the state of the world. Similarly, the advances in deep reinforcement learning techniques \citep{mnih2013playing, silver2017mastering, lillicrap2019continuous, mnih2016asynchronous} have helped make progress towards finding policies for tasks that were typically considered difficult using classical approaches for solving MDPs. One of the primary arguments for using learning-based methods is that they enable estimation of quantities required to solve a problem that would otherwise have been computationally intractable or such quantities would have even been infeasible to compute. Learning also enables agents to continually improve their decision making as they act in the environment. However, it is well known that learning-based methods have their shortcomings when it comes to generalization to previously unseen scenarios. Additionally, the application of learning-based techniques in robotics gives rise to idiosyncratic problems and research questions which are specific to robotics and hence are generally not the focus of machine learning community. That said, many limitations of learning-based methods are a shared focus of both machine learning and robotics community.

One such area of shared focus when it comes to learning is the problem of distribution shift. Learning-based methods generally involve using problem-relevant data to train algorithms to find patterns in the data and then make predictions when presented with new test data. At the heart of learning-based methods lies the assumption that the train data and the test data are drawn from the same distribution \citep{bishop2006prml}. However, since the real world is unsurprisingly messy and unpredictable, the assumption that the train data and the test data come from the same underlying distribution generally doesn't always hold for many applications. This phenomenon is known as distribution shift. Distribution shift in the data used for learning-based methods results in degraded performance in tasks that were aimed to be solved by leveraging such methods. Robotics has benefited immensely from the progress in deep learning, most notably with regards to improved perception using deep convolutional neural networks and learned planning/control with reinforcement learning techniques. However, when a robot that employs such learning-based methods comes across scenarios that are novel with respect to those that it has come across during training, the performance in the tasks that the robot was trained to do gets degraded. This problem is also becoming more prominent as a lot of robotic systems are increasingly being trained in simulation environments where the agents show superior task performance but fail to exhibit similar performance when evaluated in the real world or using real world data. This is also known as \emph{the reality gap} \citep{boeing2012leveraging}. Moreover, when a robot fails to perform its tasks as expected due to such performance degradation as a result of distribution shift, it could even potentially lead to dangerous and safety-critical consequences when deployed in the real world. Therefore, dealing with distribution shifts in robotic systems has been an important research problem.

Some of the conferences and journals where related research works in this area are generally published are International Conference on Robotics and Automation (ICRA), International Conference on Intelligent Robots and Systems (IROS), International Conference on Machine Learning (ICML), Conference on Robot Learning (CoRL), Neural Information Processing Systems (NeurIPS), AAAI Conference on Artificial Intelligence, International Conference on Computer Vision (ICCV), Conference on Computer Vision and Pattern Recognition (CVPR), European Conference on Computer Vision (ECCV), Robotics: Science and Systems Journal, International Journal of Robotics Research (IJJR).

In this paper, we discuss existing approaches that are used in robotics for decision making under distribution shift. We present a taxonomy of existing literature that discusses how robotics systems deal with distribution shifts. This paper is organized as follows. Section \ref{sec:preliminaries} briefly discusses the preliminaries required to follow the rest of the paper. Section \ref{sec:taxonomy} includes brief overview of taxonomies. Section \ref{sec:survey} surveys existing literature based on these taxonomies. Section \ref{sec:open_problems} discusses open problems. And finally, Section \ref{sec:conclusion} includes concluding remarks.

\section{Preliminaries} \label{sec:preliminaries}

\subsection{Markov Decision Process}
A general framework for sequential decision making in fully observable, stochastic environments is provided by Markov Decision Process (MDP) \citep{sutton2018reinforcement}. A Markov Decision Process consists of the following:

\begin{itemize}
    \item A set of states, $S$
    \item A set of actions, $A$
    \item State transition probabilities, $T(s', s, a)$
    \item Reward function, $R(s)$
\end{itemize}

\begin{figure}
    \centering
    \includegraphics[width=8.5cm]{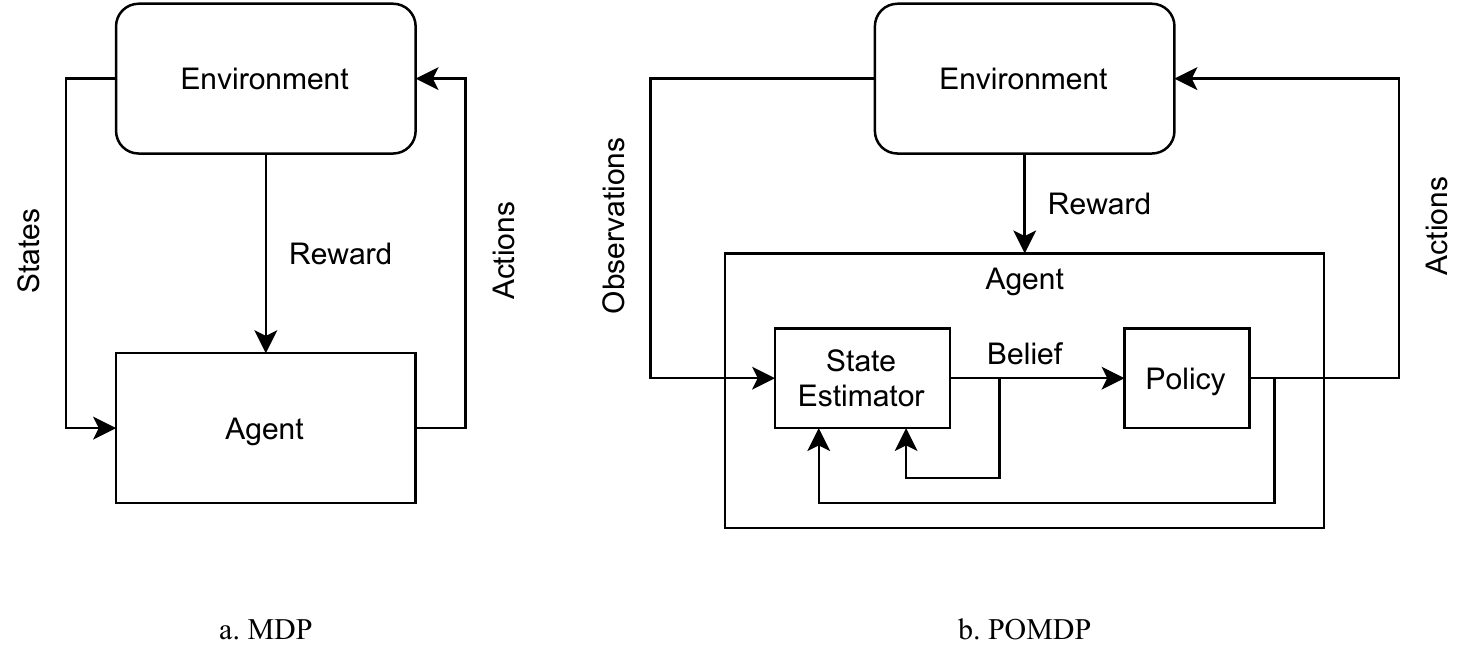}
    \caption{Illustration of agent-environment interaction in a. Markov Decision Process (MDP) and b. Partially-Observable Markov Decision Process (POMDP)}
    \label{fig:mdp}
\end{figure}

At each time step, the agent can fully observe its state and can take an action that transitions it to a new state as per the state transition probabilities with the agent receiving some reward in the process. The goal is to solve for a policy $\pi$ that specifies what the agent should do at every possible state in order to maximize the expected total reward. A policy that maximizes the total reward that an agent can accrue in expectation is called an optimal policy. Value iteration and policy iteration are two algorithms that can be used to solve for an optimal policy given an MDP. Value iteration involves alternating between policy update based on the values and value computation based on the policy. A value function for a given policy is defined over states to quantify how good it is to be in a particular state. The value of a state can be calculated using the Bellman equation as shown in Equation \ref{eq:bellman}.

\begin{equation} \label{eq:bellman}
    V^{\pi *}(s) = R(s) + \gamma \max_{a \in A(s)} \sum T(s', s, a) V^{\pi *}(s')
\end{equation}

Here, $\gamma$ denotes the discount factor and is a real number between 0 and 1 that controls how much should future rewards be prioritized over the current reward. One can also learn to represent values as a function of action in addition to state, which is known as Q-value or action-value. Value iteration assumes the equality in Equation \ref{eq:bellman} as an assignment to find optimal value function which is then used to solve for optimal policy. Policy iteration differs from value iteration in that it solves for value function at every iteration without keeping track of values from the last iteration and updates the policies using newly computed values. Depending upon how much into the future the reasoning is considered, the horizon for sequential decision making could be finite or infinite. Finite horizon problems assume that the agent has a fixed number of time steps over which it can maximize its rewards, whereas infinite horizon problems do not have a fixed number of time steps predefined and can vary or even be infinite in principle.

\subsection{Partially Observable Markov Decision Process}
Markov Decision Process assumes that the agent has complete knowledge of its state. However, in many real world problems, this assumption does not hold and the agent has to deal with state uncertainties in addition to action uncertainties. Partially Observable Markov Decision Process (POMDP) allows for dealing with state uncertainties by formulating belief over states \citep{kaelbling1998planning}. This means that the agent has to reason over every possible state it could be in at any given time making the problem more difficult to solve. Algorithms like value iteration that are used to solve an MDP become computationally intractable even for POMDPs with small state space. In the context of robotic systems, the problem of planning or navigating in unknown environments is generally formulated as a POMDP.

\subsection{Reinforcement Learning}
Although many problems can be formulated as an MDP (or a POMDP), the underlying transition probabilities and rewards may not always be directly accessible to the agent in many scenarios of interest. This means that the agent has to actually interact with the environment and learn through trial-and-error what it should do in order to maximize the return. Reinforcement learning provides a framework to solve such problems in scenarios where the agent does not have direct access to transition probabilities and/or rewards \citep{sutton2018reinforcement}. With reinforcement learning, the agent has to use observed rewards to learn a policy that maximizes the expected total reward. This can be done using two kinds of approaches: model-based approach and model-free approach.

Model-based approaches focus on first learning the transition probabilities and rewards (combinedly known as the \emph{model}) based on the agent's interaction with the environment and use this model to learn the optimal policy. The main advantage of model-based approaches is that they allow the agent to plan by reasoning about what would happen for a range of possible choices and thus make decisions based on available options. On the other hand, model-free approaches aim to solve for the optimal policy without first estimating the transition probabilities and rewards. Model-free approaches estimate value function or optimal policy from a series of interactions between the agent and the environment.

\subsection{Multi-Armed Bandit Problem}
A multi-armed bandit is a classic reinforcement learning problem in which there is a slot machine with $n$ arms (bandits) and each arm has an independent reward distribution that is hidden from the agent pulling the arms. A multi-armed bandit problem can be seen as an instance of MDP having only one state with actions that each correspond to pulling an arm from among $n$ arms. When the agent pulls an arm, it gets some reward based on the underlying reward distribution of the arm. The goal is to maximize the total reward accumulated by pulling the arms. If the agent had the knowledge of the underlying reward distributions of each arm, it could always pull the arm with the highest expected reward. However, since the underlying reward distributions are not known to the agent, a common strategy is to first identify the arm with the highest expected reward through trial and error. This requires that the agent take into consideration the exploration and exploitation trade-off while pulling the arms so as to maximize the accumulated rewards while simultaneously learning to identify the most rewarding arm.

\subsection{Autonomous Navigation and Exploration}
Many applications in robotics require robots to navigate or explore in an unknown environment on their own. The first step in this problem generally consists of building the map of the environment using data gathered from sensors like camera, LIDAR, etc. (mapping), estimating the position of the robot in the map (localization), or doing both simultaneously which is also known as simultaneous localization and mapping (SLAM) \citep{thrun2010probabilistic}. This map is then used to plan the path to reach the goal or explore the environment. Approaches for autonomous navigation/exploration has been studied widely using non-learned \citep{yamauchi1997frontier, arvanitakis2016mobile} and learned methods \citep{tai2017virtualtoreal, stein2018learning, richter2014high, kim2021plgrim, li2019deep}.

\section{Area Taxonomy} \label{sec:taxonomy}
\begin{figure*}
    \centering
    \includegraphics[width=15.8cm]{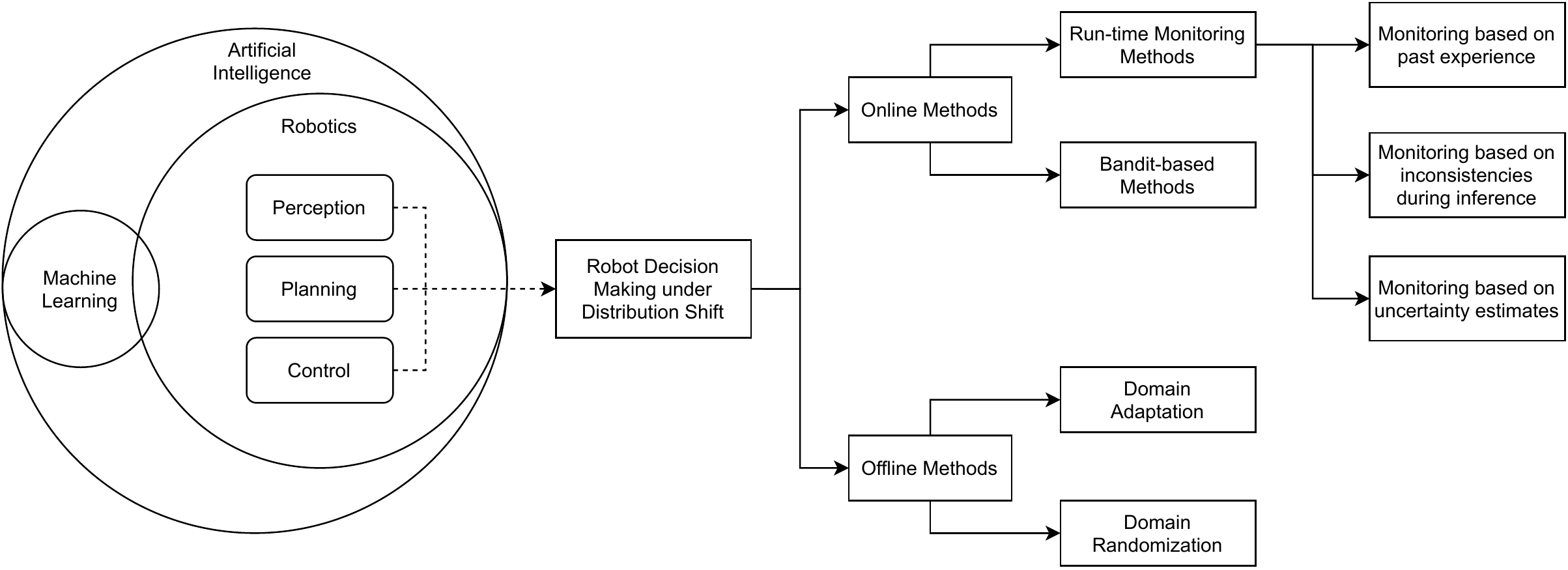}
    \caption{Overview of Taxonomy for Robot Decision Making under Distribution Shift}
    \label{fig:taxonomy}
\end{figure*}

An overview of taxonomy presented in this paper is illustrated in Figure \ref{fig:taxonomy}. Robotics and machine learning fall under the broader area of artificial intelligence. Perception, planning, and control are key components of robotic systems that help robots understand their environment, reason about how to achieve a certain task, and take actions to achieve the task respectively. Since many robotic systems leverage learning-based methods of one form or the other, decision making under distribution shift broadly relates to all three of these robotic subsystems. Distribution shifts in the state of the world or observations can be in the form of visual differences or structural differences in the real world. Visual or structural differences affect how perception systems understand and represent the state of the world. Decision making in the context of planning and control should also similarly take into account the distribution shift in the environment to avoid performance degradation or even failure of robotic systems.

Various methods have been proposed in the context of robotic systems that directly or indirectly aim to improve decision making and task performance under distribution shift. In this section, we provide a taxonomy of existing approaches in robotics literature that have emerged to tackle the problem of distribution shift. Our proposed taxonomy consists of two groups: A. Online Methods, and B. Offline Methods.

\subsection{Online Methods}
These methods aim to aid decision making during deployment through methods that are used while the robot is deployed in an environment. We divide it into two categories: 1. Run-time Monitoring Methods, and 2. Bandit-based Methods

\subsubsection{Run-time Monitoring Methods}
Run-time monitoring methods focus on monitoring various parts of the robotic systems so that when the robot is deployed in a new environment, any inferences or decisions made by the robot at run-time that might lead to potential performance degradation or failure can be detected, corrected, and avoided beforehand. These methods have been widely used in robots to monitor learning-based perception system \citep{Rahman_2021}. Run-time monitoring methods can further be divided into three sub-categories:

\begin{itemize}
    \item Monitoring based on past experience
    \item Monitoring based on inconsistencies during inference
    \item Monitoring based on uncertainty estimation
\end{itemize}

\subsubsection{Bandit-based Methods}
Bandit-based methods aim to aid decision making by formulating the problem in terms of model selection and finding the most suitable model analogous to finding the most rewarding arm in the multi-armed bandit settings. Models in this context could correspond to learned models which are each trained for decision making in a specific distribution of environment or which could each represent a specific model of object that a robotic system has to deal with for tasks like grasping or manipulation.

\subsection{Offline Methods}
These methods aim to improve decision making during deployment through methods that are implemented before the robot is actually deployed in an environment. With the advances in simulation tools and techniques, many robots are trained using simulation, and hence many offline methods have a focus on improving decision making of robots trained in simulation and deployed in real-world. Simulations provide a very accessible platform to apply various learning-based methods to train robotic systems. The difficulty and cost of gathering real world data, sample inefficiency of deep reinforcement learning algorithms, safe and faster experiments by trial-and-error in simulation, and a large number of data samples in the simulation are some of the key reasons that reinforcement learning agents are trained in simulation. However, no simulation environment can model the real world with perfect accuracy. This leads to poor performance of agents trained in simulation when they are deployed in the real world due to distribution shift in the data between simulation and the real world. It should also be noted that the offline methods for tackling distribution shifts discussed in this survey are not only limited to improving decision making for sim-to-real transfer as they can be applicable to many real-to-real or sim-to-sim applications too. We discuss some of these existing methods by grouping them in the following categories: 1. Domain Adaptation, and 2. Domain Randomization.

\subsubsection{Domain Adaptation}
Domain adaptation methods aim to handle distribution shift during deployment by training the agents with data from target domain in addition to the data from source domain. Generally, data from target domain are not easily available or the labels for these data are scarce. With domain adaptation techniques, agents are able to perform better during deployment by learning to generate features from multiple kinds of environments that are invariant to distribution shift. 

\subsubsection{Domain Randomization}
Domain randomization methods aim to generate data that covers a wide range of environments. In contrast to domain adaptation in which some data from target domain is leveraged for better adaptation to target domain, domain randomization methods aim to systematically train agents using data samples generated to account for a wide range of scenarios with the expectation that the agent will be robust to many scenarios that the agent might come across during deployment. By making use of diverse data, the training process can capture a wide range of variations in the data distribution, of which the data that the agent encounters during deployment is one of the variations.

\section{Taxonomy-based Survey} \label{sec:survey}
In this section, we present the survey of existing literature in robotics based on the taxonomy discussed in Section \ref{sec:taxonomy}.

\subsection{Online Methods}
This subsection focuses on the discussion of various online methods by dividing them into the following subcategories.
\subsubsection{Run-time Monitoring Methods}
We divide run-time monitoring methods in the existing robotics literature dividing into the following sub-categories.

\emph{i) Monitoring based on past experience}:
These methods focus on using information about how the robot had performed in the past to judge the robot's current performance and make decisions accordingly. One such example of this method is presented in \citet{guruau2018learn} where the robot's performance to detect pedestrians in outdoor environments is recorded so that it can be used again when the robot comes across the same location. Based on the past records of performance for a location, they present a probabilistic framework that either gives the robot the autonomy to drive itself or denies the autonomy and lets a human take over based evaluation using these performance records. Additionally, they also extend their approach to work for scenarios where the performance records for a location can be used for a new but visually similar location. Their approach makes use of classical decision theory to decide between when it is appropriate for the robot to take control of the driving and when the humans should intervene by specifying the associated costs of each combination of scenarios and decisions. Although their method can detect likely failure, they require humans to intervene in those cases, which might not be desirable or practical in many applications for autonomous navigation.

In \citet{daftry2016introspective}, a method to detect perception failure is proposed for autonomous navigation task in which the agent is able to evaluate the performance of its perception system. They call this ability \emph{introspection} and it allows the agent to identify any input to its perception system that is likely to lead to failure and take remedial actions accordingly. They train an introspection model that learns to predict when the input image is likely to predict a poor estimate of trajectories labels for planning and resort to predefined fail-safe maneuvers in the case of failure. They demonstrate their approach in the context of micro aerial vehicle (MAV) flight in outdoor environments. However, since their fail-safe maneuvers are hand-crafted based on the knowledge of what kind of failures are likely to occur, this might not be applicable to scenarios where one cannot intuit beforehand about what kinds of failures are likely during deployment.

In \citet{saxena2017learning}, they take the idea of \emph{introspection} from \citep{daftry2016introspective} and learn to predict potential perception failure and additionally an associated recovery maneuver based on past data about the failure and what maneuver resulted in recovery. They define likely failure scenarios as cases where the predicted trajectories are not clearly labeled as collision-free or collision-prone. By predicting the possibility of failure during autonomous quadcopter flight, it is possible to resort to simple recovery maneuvers like translate left, translate right, rotate left, and rotate right that were chosen based on domain knowledge and prevent possible collision. Although their methods extend \citep{daftry2016introspective} by additionally predicting recovery maneuver, designing the space of possible recovery maneuvers still requires some knowledge of likely failures in the context where the agent is being deployed.

\emph{ii) Monitoring based on inconsistencies during inference}:
These methods focus on detecting inconsistencies at run-time to avoid the robot making catastrophic decisions when deployed in a new environment. In \citet{mallozzi2019runtime}, they propose a method to enforce certain properties (including any safety-critical requirements) which they call \emph{invariants} that the agent has to respect all the time while exploring complex partially observable environments using reinforcement learning. Their method, called WiseML, acts as a safety envelope over any existing reinforcement learning algorithms and prevents the agents from taking actions that violate the specified invariants. WiseML makes use of a meta-monitor that evaluates the agent's possible actions to check whether they violate the invariants, and any action that is inconsistent with the invariants is not sent to the agent for execution and thus preventing the agent from taking catastrophic actions. Furthermore, their method also incorporates reward shaping to improve the agent's performance by penalizing the agent whenever the specified invariants are violated. However, since their method requires invariants to be defined based on the context, it might not be applicable in scenarios where it is not possible to clearly identify what invariants and constraints should be monitored to avoid likely failure.

In \citet{zhou2019automated}, run-time inconsistencies in semantic segmentation for autonomous driving are detected by making use of additional sensor modality. Their approach uses LIDAR as an additional sensor and use LIDAR data gathered over multiple time steps to generate point cloud which are used to generate road labels for corresponding images. These road labels are then used to validate the road labels from semantic segmentation and detect inconsistencies that might lead to failure. However, any inconsistencies due to mislabelled ground truth obtained from LIDAR data might still lead to potential inaccuracies during the validation of semantic segmentation prediction during navigation that might lead to potential failures.

\emph{iii) Monitoring based on uncertainty estimates}:
These methods focus on estimating uncertainty measures of predictions and use these estimates to detect and deal with novel scenarios that the agent comes across. In \citet{filos2020can}, they propose an epistemic uncertainty-aware planning method called \emph{robust imitative planning} that aims to detect and adapt to distribution shift by evaluating candidate plans using an ensemble of expert-likelihood models and aggregating these plans to propose a safe plan when there is a disagreement between the models. They also extend their method so that the agent can use the uncertainty estimates to query the human expert for feedback and adapt online without compromising safety.

In \citet{Richter2017SafeVN}, an auto-encoder is trained to detect out-of-distribution input to a collision-avoidance system and resort to fail-safe behavior if the input is novel. Out-of-distribution detection is done by evaluating the reconstruction error on the input image and comparing this error with the distribution of reconstruction errors for in-distribution images used for training the collision-avoidance system. Their method requires that the threshold for detecting out-of-distribution input to the collision-avoidance system be selected carefully to avoid making high false positive or false negative predictions both of which might lead to failure or significant performance degradation.

In \citet{McAllister2019RobustnessTO}, they propose a method where instead of focusing on out-of-distribution detection input to a collision prediction model, they handle out-of-distribution input by considering the effects of such input that are relevant to the collision prediction model. They do this by training a latent variable model using a variational autoencoder (VAE) that generates a distribution over a latent variable for an input image. The latent samples in this distribution are then used to sample corresponding in-distribution images which are passed to a Bayesian neural network to get an estimate of mean and variance of time-to-collision to perform risk-averse decision making. However, since VAE itself is trained using in-distribution images and used during test time for inference, their method can suffer from failure cases when inputs to variational autoencoder itself are significantly out-of-distribution.

\subsubsection{Bandit-based Methods}
Bandit-based methods make use of bandit formulation to adapt to distribution shifts. When the agent doesn't know with sufficient certainty the model of the environment it is being deployed to, one way to tackle this is to formulate a set of possible models as an analog to arms in the bandit setting and aim to identify the model that is relatively more accurate compared to the others.

In \citet{mcconachie2018estimating}, they propose to generate a set of models for deformable objects like clothes and rope and leverage the multi-armed bandit formulation to identify the best object model in the context of deformable object manipulation task. To pull an arm corresponds to using a model of the object to generate velocity commands for the gripper and execute this command and get rewards in the form of model utility based on the error in the manipulation task. Their method does not use any sensor to sense the state of the deformable object which if integrated alongside model utility could lead to better performance. 

In \citet{laskey2015multi}, they formulate the problem of identifying high-quality grasps for objects as a multi-armed bandit where each candidate grasp corresponds to an arm and each grasp is evaluated based on uncertainty in shape, pose, and friction. Their approach is able to converge to promising grasps faster than baselines. However, as the number of candidate grasps increases, it might affect the convergence rate leading to degraded performance.

\subsection{Offline Methods}
This subsection focuses on the discussion of various offline methods by dividing them into the following subcategories.

\subsubsection{Domain adaptation}
Adaptation of systems trained in one domain and tested on a different domain has been possible with recent advances in visual domain adaptation \citep{ganin2016domain, long2015learning, bousmalis2017unsupervised, hoffman2017cycada, taigman2016unsupervised}. Domain adaptation methods generally focus on training models using data in source domain while still improving performance in target domain where data may be limited or not readily available. In robotics, source domain could be a simulator where data is abundant and target domain is the real world where it may be difficult to collect a large amount of data for training. The goal is to handle the distribution shift between the data from the simulator during training and the data from real world during deployment. In addition, domain adaptation techniques have also been useful for applications beyond sim-to-real transfer and we discuss some of those below.

In \citet{bousmalis2017using}, they use simulated images that look like real images to train vision-based grasping. They propose a generative model called GraspGAN that performs pixel-level transformation on simulated images to look like real world images and train a deep convolutional network for grasping using synthesized and real world images. Their method only considers visual differences in images and no reasoning about the physical dynamics of the objects is considered which might be critical for improved performance for a wide variety of objects.

In \citet{stein2018genesis}, they propose to use CycleGAN \citep{zhu2020unpaired} to learn a mapping of images from simulation to real world which is then used to generate training images for tasks like obstacle-avoidance based flight. Their method uses labeled simulation images and unlabelled real world images to learn a generative model that can synthesize realistic-looking real world images from simulation images allowing better performance in obstacle-avoidance based flight trained using such synthesized images compared to either of those trained using simulated images or real world images alone.

In \citet{wulfmeier2017addressing}, they leverage unsupervised adversarial domain adaptation technique to address distribution shift in outdoor robotics due to changes in season and weather conditions. Their approach focuses on path segmentation tasks in which they simultaneously train a segmentation module and a discriminator module with an encoder that shares weights across both of these modules. Their method encourages the encoder to learn features that balance the importance of segmentation task and domain-invariance. However, when there are significant variations in the images used for training, the weighting for task-specific loss and adversarial loss might be more difficult resulting in poor task performance.

Similar to \citet{wulfmeier2017addressing}, the method proposed in \citet{palazzo2020domain} aims to enhance outdoor robot traversability through domain adaptation using gradient reversal technique. Their method adds a domain classification module alongside the traversability prediction module and reverses the gradients from domain classifier loss when propagating through the image encoder module. This forces the encoder to generate features that are domain-invariant and hence enable the traversability prediction module to perform better across many domains. Additionally, they also incorporate safety requirements by penalizing errors that would cause collisions, and their method is able to generalize to unseen locations.

\subsubsection{Domain randomization}

In domain randomization methods, a wide variety of environments distributions are created by randomizing properties of the environment so that the learned policies work across all of these environmental variations. The intuition with domain randomization is that by randomizing the environment properties in simulation, the agent can learn policies that work across a wide variety of environments and thereby enabling the agent to perform well when deployed in real world since the real world is expected to be one of the samples from the distribution of training environments. Randomization can be visual or at the level of environment dynamics.

In \citet{tobin2017domain}, the simulation environment is rendered with random variations to reflect the variations in the real world and an object detector is trained on these images with random camera positions, lighting conditions, object positions, and texture. This detector is then deployed in real world without any additional training and is able to perform fairly well. In \citet{peng2018sim}, they propose to learn control policy in object pushing task using a robotic arm by randomizing the dynamics of the simulator during training. By randomizing the parameters like mass of robot's links, damping of joint, height of table, mass, friction and damping of puck, etc., their method is able to learn robust policies that perform well when directly deployed in real robotic arms. In \citet{sadeghi2017cad2rl}, they propose a collision-avoidance based navigation using monocular images of randomized scenes in simulation and learn policies for indoor navigation in real world using a quadcopter. They randomize various scenes in simulation environments like wall textures, intersections, closed or open doors, lighting conditions, and orientation.

In \citet{openai2019learning}, they randomize visual appearance and physics parameters in simulation like object dimensions, object and robot link mass, surface friction, joint damping, gravity, etc. to learn control policy for dexterous in-hand manipulation of objects. The control policy is learned using object pose estimated from a separately trained pose estimator and robot fingertip locations. The training is fully done in simulation and is deployed directly in real robotic hand designed for human-level dexterity where object pose is estimated using 3 camera feeds and robot fingertip using a 3D motion capture system.
In \citet{tan2018simtoreal}, in addition to developing a more accurate actuator model and simulator latency, they randomize simulation environment parameters and add random perturbations to learn robust control policies for a quadruped robot. They randomize parameters like mass, motor friction, inertia, motor strength, latency, etc., and train control policies for trotting and galloping in simulation which can then be directly deployed in real world without further training.

Many of the domain randomization techniques enjoy the luxury of being able to randomize simulation parameters which is generally not possible in real world. Thus, these methods are limited in their approach to handling distribution shift only in sim-to-sim or sim-to-real scenarios and are not suitable for many robotic applications that require adaptation in real-to-real transfer scenarios.

\section{Discussion of Open Problems} \label{sec:open_problems}
Many open problems need to be tackled to develop techniques that enable robots to make robust and better decisions under distribution shift. This section outlines a few open problems that could serve as future directions for research in this area with a focus on applications related to autonomous navigation.

First of all, decision making under uncertainty is an open problem in itself. Any kind of uncertainty impedes the agent's abilities to make better decisions. When an agent is navigating in an unknown environment, any reasoning about the unknown environment could always turn out to be inaccurate and hence result in degraded performance or even failure to accomplish the intended tasks. In this context, distribution shift adds more complexity to decision making under uncertainty because any learned methods that the agent utilizes for making decisions under uncertainty is likely to be sub-optimal or catastrophic since the agent's assumptions about the environment do not tend to match the actual environment due to distribution shift.

Many run-time monitoring methods discussed in this survey focus on tackling distribution shift by detecting observations that might cause the perception system to behave in an undesired way. While this is important, it might also be more effective and desirable to be able to translate uncertainties in learned perception models directly into how they could affect the agent's decision making and thereby their overarching effects in the task performance of autonomous agents since task performance is generally what we ultimately care about in applications of autonomous agents.

Domain adaptation methods that aim to improve decision making under distribution shift for autonomous agents assume that some knowledge about the target domain is available even though the target domain might differ from the source domain. This means that they are limited to applications where information about target domain is given. Besides, the applicability of these methods starts to diminish as the differences in source and target domains become more prominent since a shared feature representation between source and target may be difficult to learn. This means that such methods might not be suitable for many applications where autonomous agents need to explore or navigate in environments that are not known beforehand or vary significantly among many environments where they need to be deployed. Although domain randomization techniques aim to tackle these problems by training agents with a wide variety of data distribution, they tend to be computationally expensive over a wide range of domains and also might lead to cases where tasks learned across a range of domains might have competing priorities resulting in no better performance over many environments. Besides, domain randomization techniques generally rely on access to simulator and might not be suitable for training on real world data directly. Extending domain randomization techniques to real world training settings is still an open problem.

Many approaches for planning in unknown environments discussed in this survey aim to tackle distribution shift for short-horizon planning tasks like reactive collision avoidance, short-horizon trajectory prediction, and reactive control using model-free reinforcement learning approaches. This is definitely an important problem to solve and has a wide range of useful applications. However, such methods for making decision under distribution shift might not be easily adapted to long-horizon planning tasks where the agent has to reason about the consequences of its action far into the future. Integrating model-free and model-based approaches for decision making along with capabilities that enable agents to adapt to distribution shift would be an important step towards developing agents that can achieve complicated tasks in a wide variety of environments.

There are also opportunities for bringing together online methods and offline methods for dealing with distribution shift in the context of robot decision making. For example, there are possible research directions where one could approach the adaptation of an agent to navigate in a wide variety of environments with visual or structural differences. Consider an agent that has access to multiple learned models trained in many kinds of environments. The agent can decide to use one of the learned models to plan a path towards the goal in an unknown environment (for example, the one presented in \citet{stein2018learning}). However, without the knowledge of which learned model would perform better in the environment the agent is deployed in, it would be difficult to choose between the models. One could possibly formulate this as a multi-armed bandit problem where the agent has to identify the best-performing model based on how each model performs during deployment. By monitoring and evaluating the performance of a model at run-time as analogous to getting rewards for pulling an arm in the multi-armed bandit formulation, the agent could adapt to navigate in multiple kinds of environments through online model selection. Additionally, by leveraging the correlations between performances of multiple models, it could be possible to converge to the best performing model faster than the ordinary bandit setting where the reward distribution is generally assumed to be uncorrelated among the arms. Such online model selection could also be made more efficient by integrating some form of minimal human demonstration of corrective actions that the agent can then use as a cue to identify the best-performing model for a given scenario.

\section{Conclusion} \label{sec:conclusion}
We presented a brief overview of existing literature that aims to aid or improve robot decision making under distribution shift. Although the approaches discussed in this survey contribute to making important progress towards developing agents that are able to handle distribution shifts and adapt to a wide variety of environments, there remain many open problems towards achieving that goal. We hope that the open problems that have been identified in this paper will serve as important future directions for research.

\bibliography{references.bib}

\end{document}